\documentclass[10pt,twocolumn,letterpaper]{article}

\usepackage{makecell}
\usepackage{cvpr}              

\usepackage{times}
\usepackage{epsfig}
\usepackage{graphicx}
\usepackage{amsmath}
\usepackage{amssymb}

\usepackage{multirow}
\usepackage{tabularx}
\usepackage{booktabs}
\usepackage{longtable}
\usepackage{subcaption}
\usepackage{colortbl}

\usepackage[accsupp]{axessibility}  
\usepackage[font=footnotesize,labelfont=bf]{caption}
\usepackage[belowskip=-5pt,aboveskip=3pt]{caption} 

\usepackage{dsfont}
\usepackage{pifont}
\usepackage{url}
\usepackage{xcolor}
\usepackage{amsthm}
\usepackage{adjustbox}
\usepackage{comment}
\usepackage{algorithm} 

\usepackage[utf8]{inputenc} 
\usepackage[T1]{fontenc}    
\usepackage{url}            
\usepackage{amsfonts}       
\usepackage{nicefrac}       
\usepackage{microtype}      
\usepackage{enumitem}
\usepackage{blindtext}
\usepackage{sidecap}
\usepackage{wrapfig}

\definecolor{citecolor}{RGB}{65,105,225}

\usepackage{dsfont}

\definecolor{dg}{rgb}{0,0.694,0.298}
\definecolor{purple}{rgb}{0.4,0.176,0.569}
\definecolor{royalblue}{RGB}{65,105,225}
\usepackage{pifont}
\newcommand{\figref}[1]{Fig.~\ref{#1}}
\newcommand{\reqref}[1]{Eq.~(\ref{#1})}
\newcommand{\secref}[1]{Sec.~\ref{#1}}
\newcommand{\tableref}[1]{Table~\ref{#1}}

\makeatletter
\DeclareRobustCommand\onedot{\futurelet\@let@token\@onedot}
\def\@onedot{\ifx\@let@token.\else.\null\fi\xspace}
\def\eg{\emph{e.g}\onedot} 
\def\ie{\emph{i.e}\onedot}

\def\etal{\emph{et al}\onedot}
\makeatother

\definecolor{americanrose}{rgb}{1.0, 0.01, 0.24}

\theoremstyle{plain}
\theoremstyle{definition}

\theoremstyle{remark}

\usepackage[textsize=tiny]{todonotes}

\definecolor{cvprblue}{rgb}{0.21,0.49,0.74}
\usepackage[pagebackref,breaklinks,colorlinks,allcolors=cvprblue]{hyperref}

\def\paperID{9807} 
\def\confName{CVPR}
\def\confYear{2025}

\title{\textsc{TruePose:} Human-Parsing-guided Attention Diffusion for \\ Full-ID Preserving Pose Transfer}

\author{
\textbf{Zhihong Xu}\textsuperscript{1}
\ 
\textbf{Dongxia Wang}\textsuperscript{1,*}
\ 
\textbf{Peng Du}\textsuperscript{2}
\ 
\textbf{Yang Cao}\textsuperscript{2}
\ 
\textbf{Qing Guo}\textsuperscript{3,4,*}
\\
\textsuperscript{1} Zhejiang University
\quad
\textsuperscript{2} Alibaba Group
\\
\quad
\textsuperscript{3} 
Institute of High Performance Computing (IHPC), A*STAR, Singapore
\\
\quad
\textsuperscript{4} 
Centre for Frontier AI Research (CFAR), A*STAR, Singapore
}

\begin{document}
\maketitle 

\if TT\insert\footins{\noindent\footnotesize{
*Corresponding authors}}\fi

\begin{abstract}
Pose-Guided Person Image Synthesis (PGPIS) generates images that maintain a subject's identity from a source image while adopting a specified target pose (e.g., skeleton). While diffusion-based PGPIS methods effectively preserve facial features during pose transformation, they often struggle to accurately maintain clothing details from the source image throughout the diffusion process. 
This limitation becomes particularly problematic when there is a substantial difference between the source and target poses, significantly impacting PGPIS applications in the fashion industry where clothing style preservation is crucial for copyright protection.
Our analysis reveals that this limitation primarily stems from the conditional diffusion model's attention modules failing to adequately capture and preserve clothing patterns.
To address this limitation, we propose human-parsing-guided attention diffusion, a novel approach that effectively preserves both facial and clothing appearance while generating high-quality results.
We propose a human-parsing-aware Siamese network that consists of three key components: dual identical UNets (TargetNet for diffusion denoising and SourceNet for source image embedding extraction), a human-parsing-guided fusion attention (HPFA), and a CLIP-guided attention alignment (CAA). The HPFA and CAA modules can embed the face and clothes patterns into the target image generation adaptively and effectively.
Extensive experiments on both the in-shop clothes retrieval benchmark and the latest in-the-wild human editing dataset demonstrate our method's significant advantages over 13 baseline approaches for preserving both facial and clothes appearance in the source image.
\end{abstract}

\section{Introduction}
\label{sec:intro}

\begin{figure*}[t]
    \centering
    \includegraphics[width=1.0\linewidth]{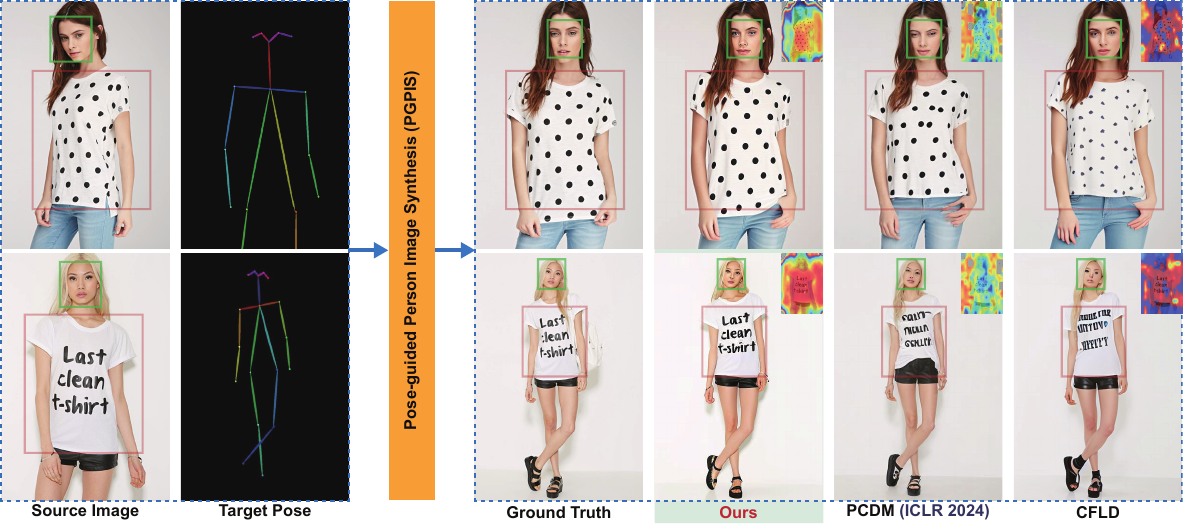}
    \caption{Pose-guided person image synthesis (PGPIS) task and comparison among CFLD  \cite{lu2024coarse}, PCDM (ICLR'24) \cite{shen2024advancing}, and our methods. The two SOTA methods fail to preserve the clothing patterns and textures. The main reason is that the image encoder overlooks the clothing details (See the ``Feature Attention Map (Feat. Att. Map)").  In contrast, our method can generate high-quality images with preserved face and clothing patterns. 
    }
    \label{fig:fig_intro}
\end{figure*}

Images of well-dressed individuals are widely used in shopping stores for advertising. To achieve a high-quality image, a model must wear specific clothing and adopt a designated pose. Once these images are captured and deployed, it becomes difficult for stores to change the model's pose.
Otherwise, they must invite the model to wear the same clothes and undergo the photography process again, which is time-consuming and costly.
The well-defined task, \ie, pose-guided person image synthesis (PGPIS), could easily meet the above requirement efficiently. 
PGPIS aims to generate an image that matches the specified pose condition (\eg, skeleton) while retaining the same appearance (person and clothing) as the source image (See the inputs in \figref{fig:fig_intro}) \cite{ma2017pose,liu2019liquid,zhu2019progressive}.
Such a task is challenging due to the potential significant discrepancy between the source and target poses.

Previous works formulate the task as the deep generation problem and employ generative adversarial network (GAN) \cite{zhou2022cross,zhang2022exploring,zhang2021pise,sarkar2021style,tang2020xinggan,zhu2019progressive,siarohin2018deformable} and variational autoencoder (VAE) \cite{esser2018variational} to achieve the goal. 
However, GAN-based methods struggle with unstable training and generating high-quality images. VAE-based methods usually suffer from some artifacts like blur. 

Recently, diffusion models have demonstrated powerful capabilities in generating high-quality images \cite{ho2020denoising,song2020score} with flexible control conditions \cite{rombach2022high}.
Researchers have developed diffusion-based PGPIS methods, achieving impressive results \cite{bhunia2023person,lu2024coarse,shen2024advancing}.
In particular, the state-of-the-art (SOTA) methods, \eg, coarse-to-fine latent diffusion (CFLD) \cite{lu2024coarse} and progressive conditional diffusion (PCDM) \cite{shen2024advancing}, can generate realistic images with preserved human pose and face ID (See in \figref{fig:fig_intro}).
However, they struggle to transfer clothing patterns and textures (\ie, clothing ID) from the source image to the target image.
%
%
We display four cases in \figref{fig:fig_intro} and observe that: \ding{182} CFLD and PCDM can hardly preserve clothing patterns and textures. The two methods fail to transfer the texts ($1^\text{st}$ and $4^\text{th}$ cases), regular texture ($2^\text{nd}$ case), and irregular textures ($3^\text{rd}$ case).
%
%
\ding{183} As the discrepancy between the source and target poses increases, it becomes more challenging to preserve the clothing IDs. For example, with similar source images ($1^\text{st}$ case vs. $4^\text{th}$ case), PCDM can reproduce some words in the $1^\text{st}$ case but loses all words in the $4^\text{th}$ case.
\ding{184} We calculate the feature attention map of baseline methods for each case, which shows that both methods pay less attention to the clothing regions. We will conduct a statistical analysis in \secref{sec:limitation} for further discussion.

These observations inspire us to design a novel PGPIS framework that preserves both facial and clothing patterns. Our approach leverages the person-parsing map of the source image to guide the encoding process, ensuring the features focus on both face and clothing regions.
We propose a human-parsing-aware Siamese network with three main modules: \textit{First}, we design a dual idenitical UNets (\ie, SourceNet and TargetNet). TargetNet is used for diffusion denoising and SourceNet is designed for extracting source image embedding.
\textit{Second}, we introduce the human-parsing-guided fusion attention, which corrects the attention of TargetNet's embedding on the facial and clothes regions according to the guidance of the source parsing map and source image embeddings.
\textit{Third}, we propose CLIP-guided attention alignment to further refine the corrected TargetNet embedding based on the consistent constraints across different semantic regions.
The proposed modules could be inserted into different layers of the TargetNet.
We validate our method on the in-shop clothes retrieval benchmark with two resolutions ($512\times352$, $256\times176$) and compare with 13 SOTA baselines.
As shown in \figref{fig:fig_intro}, our method can generate high-quality images with well-preserved facial appearance and clothing patterns.

\section{Related Work}
\label{sec:related_work}

\textbf{Pose-guided person image synthesis.}
PGPIS achieves pose transformation in a complex spatial context while maintaining consistency with the detailed features of the source image, which was proposed by Ma \etal  \cite{ma2017pose}. 
Early approaches \cite{esser2018variational,ma2017pose} employ conditional generative adversarial networks (CGANs)  \cite{goodfellow2014generative, mirza2014conditional} to guide the generation of target images using the source image as conditional information \cite{ma2017pose,men2020controllable}. However, due to the significant spatial information differences between the source image and the target image, as well as the sparse guidance from the skeleton map, directly achieving pose transfer is highly challenging. To address the challenges, some works decouple and separately optimize pose and appearance information\cite{ma2018disentangled}, or use attention mechanisms to better establish the mapping relationship between pose and appearance\cite{ren2022neural,zhang2022exploring,zhou2021cocosnet}. 
On the other hands, some approaches introduce more detailed and dense supplementary information, such as UV \cite{sarkar2021style}, dense-pose \cite{han2023controllable}, and parsing maps \cite{men2020controllable,zhang2021pise,zhou2022cross,lv2021learning}, to alleviate inconsistencies between poses. 
Recent approaches adopt diffusion as the generative model\cite{bhunia2023person,shen2024advancing,lu2024coarse,han2023controllable}. 
By incorporating conditional information into iterative forward-backward denoising procedures \cite{ho2020denoising}, diffusion-based methods outperform GAN-based approaches in terms of both image generation quality and pose control effectiveness.
PIDM is the pioneering endeavor to integrate the diffusion model into PGPIS, which designs a conditional texture diffusion model and a disentangle classifier-free guidance mechanism \cite{ho2022classifier}. 
PoCoLD and PCDM aim to achieve pose alignment through a pose-constrained attention mechanism and progressive multi-stage refinement, respectively \cite{han2023controllable,shen2024advancing}.
CFLD \cite{lu2024coarse} endeavors to attain coarse-to-fine generation by extracting coarse prompt information from the source image, followed by refinement through a learnable Perception-Refined Decoder and Hybrid-Granularity Attention to achieve fine-grained results. 

\textbf{Conditional diffusion models.}
The significant potential of diffusion models in traditional unconditional generative tasks has recently been evidenced\cite{ho2020denoising,song2020denoising,song2020score}. 
In contrast to conventional single-step generative models such as Generative Adversarial Networks (GANs)\cite{mirza2014conditional} and Variational Autoencoders (VAEs)\cite{kingma2013auto}, diffusion relies on multi-step denoising sampling of initial noise to generate high-quality data, thereby enabling the production of detailed and diverse outputs. The advancement of various conditional diffusion models further enhances its practicality and versatility. Classifier-free methodologies devise implicit classifiers to guide the weights of controllable and uncontrollable processes, thus facilitating conditional generation\cite{ho2022classifier}. Furthermore, Latent Diffusion Model(LDM)\cite{rombach2022high} utilize denoising process by encoding original images into a low-dimensional latent space and integrate cross-attention mechanisms to introduce multi-modal conditions, thereby significantly expanding the diversity of control information. 
Prior LDM-based approaches\cite{han2023controllable,shen2024advancing,lu2024coarse} rely on pre-trained encoders to extract appearance information from the source image. 
We argue that such encoders are ineffective in capturing fine-grained details, which hinders the generation of complex images. In contrast, our proposed framework utilizes guided attention to capture complex detail features, improving the preservation of full identities and ensuring person consistency.
Our extensive experimental analysis confirms the limitations of traditional approaches. (\secref{sec:limitation}).

\section{Preliminaries and Limitations}
\label{sec:preliminary}


\subsection{Diffusion-based PGPIS }
\label{subsec:diff_pgpis}

Given a source image $\mathbf{I}_s$ containing a person with pose $\mathbf{p}_s$ (represented as a skeleton map) and a target pose $\mathbf{p}_\tau$, PGPIS aims to generate a target image $\mathbf{I}_\tau$ that preserves the person's appearance (should contain both facial and clothing patterns) while adopting the target pose.
The SOTA diffusion-based PGPIS methods are designed based on the stable diffusion (SD) \cite{rombach2022high} that involves a forward diffusion process and a backward denoising process of $T$ steps.
%
%
The forward diffusion process progressively add random Gaussian noise $\epsilon\in \mathcal{N}(0,\mathbf{I})$ to the initial latent $\mathbf{z}_0$. At the $t$th timestep, we can formulate it as
\begin{align} \label{eq:forward_diffusion}
    \mathbf{z}_t = \sqrt{\bar{\alpha}_t}\mathbf{z}_0+ \sqrt{1-\bar{\alpha}_t}\epsilon, t\in[1,T]
\end{align}
where $\bar{\alpha}_1,\bar{\alpha}_2,\ldots,\bar{\alpha}_T$ are calculated from a fixed variance schedule. The denoising process uses an UNet to predict the noise, \ie, $\epsilon_\theta(\mathbf{z}_t, t, \mathcal{C},\mathcal{O}_\beta)$ and remove the noise from the latent.
The set $\mathcal{C}$ involves related conditions and the function $\mathcal{O}_\beta(\cdot)$ defines the way of embedding condition features into the UNet. 
For PGPIS task, we can set  $\mathcal{C} =\{\mathcal{X}_\text{s},\mathcal{X}_\text{tp}, \mathcal{X}_\text{sp}\}$ where $\mathcal{X}_\text{s}$ denotes the features of source image and $\mathcal{X}_\text{tp}$ and $\mathcal{X}_\text{sp}$ contains the features of target pose and source pose.
To train the UNet $\epsilon_\theta(\cdot)$, the predicted noise should be the same as the sampled noise during the forward process
\begin{align} \label{eq:loss_diffusion}
    \mathcal{L}_\text{mse} = \mathds{E}_{\mathbf{z}_0, \mathcal{C}, \epsilon, t}(\|\epsilon-\epsilon_\theta(\mathbf{z}_t, t, \mathcal{C},\mathcal{O}_\beta)\|^2_2).
\end{align}
During the inference with trained $\epsilon_\theta$ and $\mathcal{O}_\beta$, we use image encoder and pose encoder to extract features of the inputs (\ie, $\mathbf{I}_\text{s}$, $\mathbf{p}_\text{s}$, $\mathbf{p}_\text{t}$) and get $\mathcal{C} =\{\mathcal{X}_\text{s},\mathcal{X}_\text{tp},\mathcal{X}_\text{sp}\}$.
Then, given the latent $\mathbf{z}_T$, we perform the backward denoising process iteratively to generate the predicted target image. 

\textbf{Coarse-to-fine latent Diffusion (CFLD) \cite{lu2024coarse}.} 
The SOTA method CFLD utilizes the pre-trained Swin-B \cite{liu2021swin} to extract multi-layer features (\ie, $\mathcal{X}_\text{s}$) of the source image and the Adapter to extract multi-layer features \cite{mou2023t2i} (\ie, $\mathcal{X}_\text{tp}$) of the target pose. Moreover, CFLD proposes two new modules as the $\mathcal{O}_\beta$ to embed the condition features effectively.

\textbf{Progressive conditional Diffusion models (PCDM) \cite{shen2024advancing}.} Shen \etal proposed three-stage diffusion models to progressively generate high-quality synthesized images. The first stage is to predict the global embedding of the target image; the second stage is to get a coarse target estimation; the final stage is to refine the coarse result. Each stage relies on a diffusion model that could be approximately formulated with \reqref{eq:forward_diffusion} and \reqref{eq:loss_diffusion} with different condition setups. In particular, all three stages use the pre-trained image encoders, \ie, CLIP and DINOv2.

\subsection{Empirical Study}
\label{sec:limitation}

\begin{SCfigure}{}{}
    \includegraphics[width=0.75\linewidth]{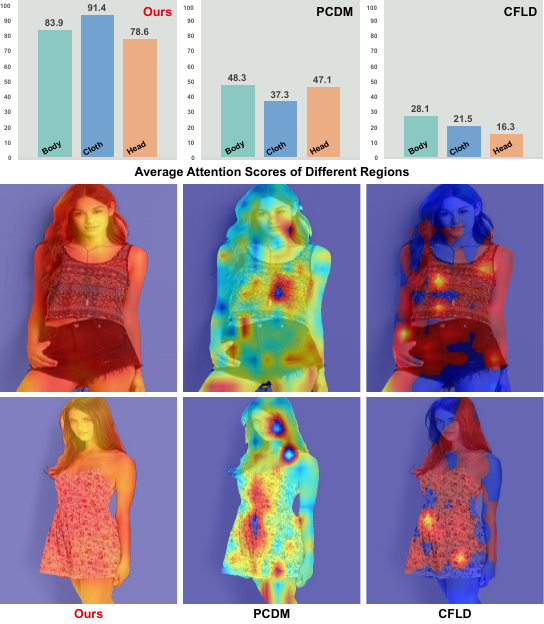}
    \caption{\textbf{Top:} comparing average attention scores of different regions within source images. \textbf{Bottom:} visualization results of attention maps of our method and two baseline methods.
    }
    \vspace{-15pt}
\label{fig:limitation}
\end{SCfigure}

As shown in \figref{fig:fig_intro}, the state-of-the-art methods CFLD \cite{lu2024coarse} and PCDM \cite{shen2024advancing} can hardly preserve the clothing patterns and textures.
The feature attention maps in \figref{fig:fig_intro} demonstrate a superficial reason, that is, the SOTA methods' features did not focus on the clothing. 
To further validate this observation, we perform a statistical analysis.
Specifically, we randomly collect 50 examples from the in-shop clothes retrieval benchmark \cite{liuLQWTcvpr16DeepFashion} and use the two SOTA methods to handle these examples.
For each sample, we calculate the feature attention map and get its parsing map.
Then, we can count the attention values at different parsing regions.
After that, we can obtain the average attention values within different regions among all examples.
For CFLD, we obtain the feature attention map by calculating the final layer's average value of the softmax layer in the Swin-B. 
For PCDM, we follow the steps outlined in the open-source code of PCDM to calculate the attention map.
As the results are shown in \figref{fig:limitation}, PCDM using pre-trained CLIP and DINOv2 feature pay less attention to the clothes when we compare the attention values on clothes with the ones on the other two regions.
CFLD uses pre-trained Swin-B features, paying less attention to both cloth and head regions.
The observation inspires us to design a novel framework, which should encode the source image with both face and cloth key information preserved.
Our intuition is to leverage person-parsing maps to guide the image encoding process and inject the parsing-guided embeddings into diffusion generation.

\section{Human-Parsing-aware Siamese Network}
\label{sec:method}

\begin{figure*}[t]
    \centering
    \includegraphics[width=1.0\textwidth]{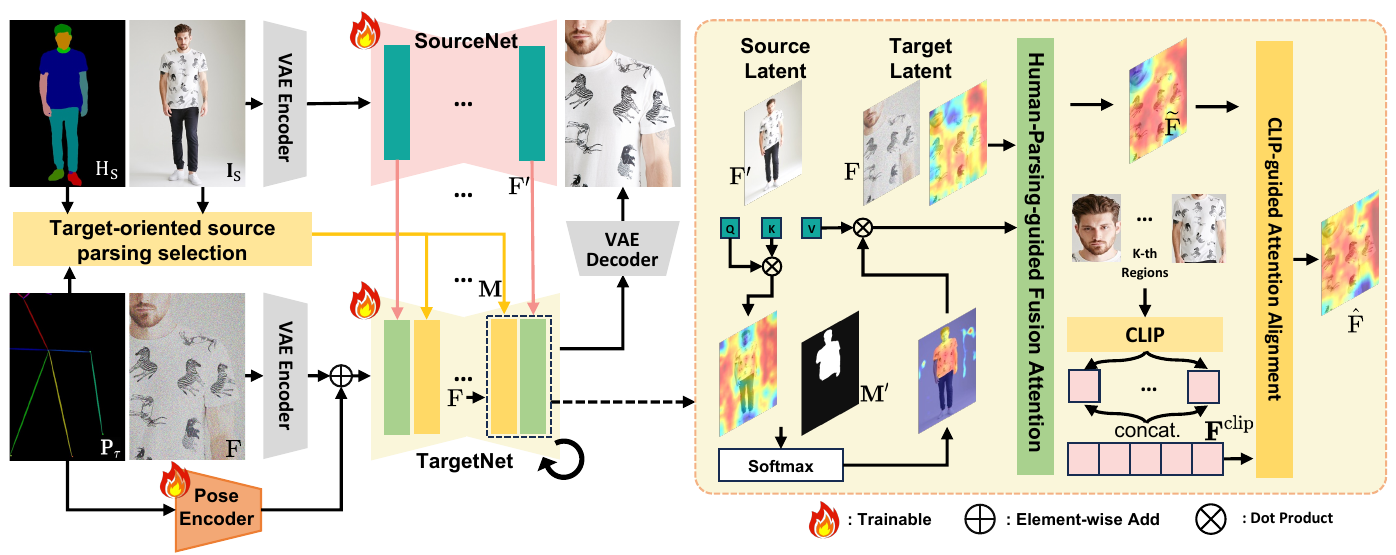}
    \caption{Pipeline of the proposed human-parsing-guided attention diffusion model.}
    \label{fig:pipeline}
\end{figure*}


\subsection{Overview}
\label{subsec: overview}

Following the task definition of PGPIS, we have similar inputs including source image $\mathbf{I}_\text{s}$ and target pose $\mathbf{p}_\tau$.
We begin by acquiring person-parsing maps. For the source image $\mathbf{I}_\text{s}$, we utilize the person-parsing approach of \cite{li2020self} to generate parsing map $\mathbf{H}_\text{s}$, which semantically segments the human body into categories such as arm, leg, dress, and skirt.

With the parsing map and inputs, we also use the diffusion model to generate the target image and the key problem is how to set the condition $\mathcal{C}$ and the fusion function $\mathcal{O}_\beta$ in \reqref{eq:loss_diffusion}. Different from previous works using pre-trained encoders to extract embeddings of source, pose, and parsing images, we design a human-parsing-aware Siamese network. 

\textit{First}, we set an encoder denoted as $\phi(\cdot)$ to extract the latent of source images and the noisy target image during the diffusion process, which can be formulated as $\mathbf{z}_\text{s} = \phi(\mathbf{I}_\text{s})$.

\textit{Second}, we build a Siamese network containing two UNets denoted as TargetNet (\ie, $\epsilon_\theta$) and SourceNet (\ie, $\epsilon_{\theta'}$).
TargetNet is the UNet in the stable diffusion, taking the noisy latent as input $\mathbf{z}_t$, predicting the noise amount at $t$th timestamp for inverse denoising. 
SourceNet has the same architecture as the TargetNet and takes $\mathbf{z}_\text{s}=\phi(\mathbf{I}_\text{s})$ as the input and output $L$ embeddings from $L$ layers of $\epsilon_{\theta'}$,
\begin{align} \label{eq:sourcenet}
    \mathcal{F}' = \{\mathbf{F}_l'\}_{l=1}^L = \epsilon_{\theta'}(\mathbf{z}_\text{s}).
\end{align}
Meanwhile, we leverage the parsing map to split the source image into $M$ regions corresponding to their $M$ categories and get $\{\mathbf{R}_i\}_{i=1}^M$. Each region $\mathbf{R}_i$ is a rectangle region wrapping the $i$th category. Then, we use the pre-trained CLIP encoder to extract the embeddings of all regions
\begin{align} \label{eq:clip}
    \mathcal{F}_\text{clip}= \{\mathbf{F}_i^\text{clip}\}_{i=1}^M = \{\text{CLIP}(\mathbf{R}_i)\}_{i=1}^M.
\end{align}

We leverage SourceNet and CLIP encoder embeddings of the source image as conditions in our diffusion. The design is motivated by two key advantages: \ding{182} The SourceNet, sharing identical architecture and initial weights with the TargetNet, naturally produces embeddings that are well-aligned with TargetNet's feature representations across all layers. This alignment facilitates effective transfer of source image information during the diffusion process. \ding{183} The CLIP encoder provides rich semantic embeddings that effectively capture region-specific features, enhancing semantic consistency across different areas of the generated image. 
    
We formulate the noise prediction for diffusion by 
%
\begin{align}\label{eq:targetnet}
    \hat{\epsilon}_t = \epsilon_\theta(\mathbf{z}_t, t, \mathcal{C},\mathcal{O}_\beta), 
    \text{s.t.}, \mathcal{C}=\{\mathcal{F}', \mathcal{F}_\text{clip}\},
\end{align}
where $\hat{\epsilon}_t$ is the predicted noise at $t$th timestamp. The function $\mathcal{O}_{\beta}$ is the way to fuse conditions $\mathcal{C}$ into the TargetNet, and the key problem is how to design the function $\mathcal{O}_{\beta}$. 

We propose a two-stage fusion strategy with the above two conditions, respectively.
In \secref{subsec:fusion}, we introduce human-parsing-guided fusion attention that enhances the embedding's attention on the body, clothes, and facial regions in the target image according to the source parsing map and target pose. 
However, the output embedding still presents low scores in some regions and different source images have different lower-score regions. To fix this problem, we further propose a CLIP-guided attention alignment in \secref{subsec:alignment} that can enhance the regions with lower attention scores automatically.

\subsection{Human-Parsing-guided Fusion Attention}
\label{subsec:fusion}

At the $l$th layer of the Siamese network, we obtain embeddings from both the TargetNet and SourceNet, denoted as $\mathbf{F}_l$ and $\mathbf{F}_l'$ respectively.
Our method generates a refined embedding $\tilde{\mathbf{F}}_l$ by fusing $\mathbf{F}_l$ with $\mathbf{F}_l'$.
For notational simplicity, we will omit the layer index $l$ in subsequent discussions.

\textbf{Target-oriented source parsing selection.} The given target pose $\mathbf{p}_\tau$ contains the coordinates of the skeleton points and the categories of all points. We denote the category set of the target pose as $\mathcal{A}_{\tau}=\{a_j^\tau\}$. 
Meanwhile, we have the category set of the parsing map $\mathbf{H}_\text{s}$ and denote it as $\mathcal{A}_{\text{s}}=\{a_j^\text{s}\}$. Each category $a_j^\text{s} \in \mathcal{A}_{\text{s}}$ corresponds to a mask indicating the category's region in the source image and is denoted as $\mathbf{M}_{a_j^\text{s}}$.
Then, we can get a mask map $\mathbf{M}$ that is the combination of all categories contained in the target pose, that is, we have $\mathbf{M}=\bigcup_{a} \{\mathbf{M}_a|a\in \mathcal{A}_{\tau}\}$.

\textbf{Selected-parsing-reweighed attention.} 
Previous works in pose-guided generation methods \cite{hu2024animate,xu2023magicanimate} have demonstrated that modifying self-attention can help regularize the identity of images.
They normally concatenate $\mathbf{F}$ and $\mathbf{F}'$ and input them into the self-attention layers of TargetNet for fusion and then decompose. We argue that such a fusion method is not suitable for tasks involving significant pose transformations, as it can lead to distortion of image generation(more details in \secref{subsec: ablation study}).
This limitation arises from the inability to effectively guide the model's focus toward the interest region. In response, we propose human-parsing-guided fusion attention, which leverages binary mask $\mathbf{M}$ to reweight the embeddings of $\mathbf{F}'$.

Specifically, we resize the $\mathbf{M}$ using interpolation to match the corresponding size of $\mathbf{F}'\in \mathds{R}^{h\times w \times c}$. 
Then, we calculate the query, key, and value of $\mathbf{F}'$ through $({\mathbf{Q}', \mathbf{K}', \mathbf{V}'}) = ({\mathbf{W}^{Q'} \mathbf{F}'},{\mathbf{W}^{K'} \mathbf{F}'},{\mathbf{W}^{V'} \mathbf{F}'})$, and get the attention map $\mathbf{A}' ={ (\textbf{Q}_l' \textbf{K}_l'^\top)} / {\sqrt{d}}$ that indicates the focusing regions within $\mathbf{F}'$.
Then, we derive the mask weight matrix $\textbf{M}'$ based on the attention map $\mathbf{A}'$: 
\begin{align} \label{eq:attreweight}
    \mathbf{M}'_{i, j} = \begin{cases}
     1 + \delta & \text{if } \mathbf{A}_{i,j}'>0 \text{ and } \mathbf{M}_{i,j}=1 \\
     \delta & \text{if } \mathbf{A}_{i,j}'<0 \text{ and } \mathbf{M}_{i,j}=1 \\
     \sigma & \text{if } \mathbf{A}_{i,j}'>0 \text{ and } \mathbf{M}_{i,j}=0 \\
     1 + \sigma & \text{if } \mathbf{A}_{i,j}'<0 \text{ and } \mathbf{M}_{i,j}=0 \\
 \end{cases}  \\ \nonumber
     \text{s.t.}, i \in \{0,\ldots, h \}, j \in \{0, \dots, w \},
\end{align}
where $\sigma>0$ and $\delta>0$ are hyperparameters. Intuitively, the \reqref{eq:attreweight} assigns higher weights to the embeddings with higher attention through $1+\delta$ and  $1+\sigma$ for the masked region and unmasked region, respectively. In our experiments, we empirically set $\sigma$ and $\delta$ as 0.3 and 0.6 to avoid overfitting and distortion of the background.
Then, we compute the hidden states via reweighted attention :
\begin{align}\label{eq:reweightedatt}
     \mathbf{H}' & = \text{RwSelfAtt}(\mathbf{Q}',\mathbf{K}',\mathbf{V}',\mathbf{M}'),  \\  
     & = \text{SoftMax}(({\textbf{Q}' {\textbf{K}'}^{\top}}/{\sqrt{d}}) \odot \mathbf{M}' ) \textbf{V}'. \nonumber
\end{align}
We further fuse $\mathbf{H}'$ and $\mathbf{F}$ via cross attention:
\begin{align} \label{eq:fuseatt}
    \tilde{\mathbf{F}} & = \text{CrossAtt}(\mathbf{Q},\mathbf{K}_h',\mathbf{V}_h'),  \\  
     & = \text{SoftMax}(\mathbf{Q} {\mathbf{K}_h'}^{\top}/{\sqrt{d}}) \mathbf{V}_h', \nonumber
\end{align}
where $({\textbf{Q}, \textbf{K}_h', \textbf{V}_h'}) = ({\mathbf{W}^{Q} \mathbf{F}},{\mathbf{W}^{K_h'} \mathbf{H}'}, {\mathbf{W}^{V_h'} \mathbf{H}'})$.
Then, we can compare the attention map of the original embedding $\mathbf{F}_l$ and the one of the new embedding $\tilde{\mathbf{F}}$. As shown in \figref{fig:pipeline}, we find that the model focuses more on the areas indicated by the mask during the diffusion process, allowing for better feature extraction and effective pose transfer.
%
We can insert the proposed attention module into different layers.

\subsection{CLIP-guided Attention Alignment}
\label{subsec:alignment}

Human-parsing-guided fusion attention facilitates the correction of the embedding $\mathbf{F}$ to highlight regions that would appear in the target image, but some regions still have low attention scores.
To address this limitation, we propose leveraging CLIP embeddings of local image regions to further refine the attention in $\tilde{\mathbf{F}}$.

We first split the attention map $\mathbf{A}'$ into $M$ regions according to the parsing map $\mathbf{H}_\text{s}$ and calculate the average attention scores of each parsing region. Then, we select $K$ regions with the lowest average attention scores. We have the $K$ selected CLIP embeddings $\{\mathbf{F}^\text{clip}_i\}_{i=1}^K$ via \reqref{eq:clip}, concatenate them, and get the embedding $\mathbf{F}^\text{clip}$.
We perform the cross-attention with $\tilde{\mathbf{F}}$ and output the final refined embedding
\begin{align}
    \hat{\mathbf{F}} & = \text{CrossAtt}(\tilde{\mathbf{Q}},\mathbf{K}^\text{clip},\mathbf{V}^\text{clip}), \\ \nonumber
    & = \text{SoftMax}(\tilde{\mathbf{Q}} {\mathbf{K}^\text{clip}}^{\top}/{\sqrt{d}}) \mathbf{V}^\text{clip},
\end{align} 
where $\tilde{\mathbf{Q}}, \mathbf{K}^\text{clip}, \mathbf{V}^\text{clip}= 
(\mathbf{W}^{\tilde{Q}}\tilde{\mathbf{F}}, \mathbf{W}^{K^\text{clip}}\mathbf{F}^\text{clip}, \mathbf{W}^{V^\text{clip}}\mathbf{F}^\text{clip})$. The emebdding $\tilde{\mathbf{F}}$ is calculated by \secref{subsec:fusion} with \reqref{eq:fuseatt}.
It can be observed that while $\tilde{\mathbf{F}}$ has extracted the majority of fine-grained features, some areas, particularly the face, remain underrepresented. A comparison between $\tilde{\mathbf{F}}$ and $\hat{\mathbf{F}}$ reveals that our use of CLIP embeddings effectively enhances features in these local regions.

We can insert the proposed attention module and the alignment module into different layers of the TargetNet and enhance the embeddings across multiple layers.

\subsection{Implementation Details}
\label{subsec:implementation}

\textbf{Optimization and sampling.}
During the training process, we also adopt classifier-free guidance \cite{ho2022classifier}, which is a strategy widely used in diffusion models to enhance the quality and control of the generated images. To achieve that we set conditions $\mathcal{C}$ to 0 with a random probability of $\eta$\%.
The final training loss function is rewritten as
\begin{align} 
    \label{eq:saimese_loss}
      \mathcal{L} = \mathds{E}_{\mathbf{z}_0, \mathcal{C}, \epsilon, t}(\|\epsilon-\epsilon_\theta(\mathbf{z}_t, t, \mathcal{C},\mathcal{O}_\beta)\|^2_2).
\end{align}
%
%
During inference, given a randomly initialized Gaussian noise, we utilize TargetNet (\ie, $\epsilon_\theta(\cdot)$) to predict the noise at the $t$th timestep and employ classifier scale $\omega$ to regulate the strength of guidance. The sampling formula is as follows:
\begin{equation}
    \begin{split}
    \hat{\epsilon}_{t}=&\epsilon_{\theta}(z_t,t,\varnothing,\mathcal{O}_{\beta}) + \\
    & \omega \cdot(\epsilon_{\theta} (\mathbf{z}_t,t,\mathcal{C},\mathcal{O}_{\beta})-\epsilon_{\theta}(z_t,t,\varnothing,\mathcal{O}_{\beta})),
    \end{split}
\end{equation}
Target latents can be obtained through multi-step denoising process, and mapping them back to the original image space we can acquire the target images. 

\noindent \textbf{Model details.}
Our framework is derived by modifying the structure and weights of Stable Diffusion v1.5, based on the Hugging Face Diffusion library. PoseEncoder is a lightweight network with four convolutional layers. 
For training, we run on 4 NVIDIA A800 GPUs, with a batch size of 12 for $512\times 352$ and batch size of 60 for $256\times 176$. The training process consists of 50 epochs and uses the AdamW optimizer with a fixed learning rate of $1e^{-5}$. Classifer-free dropout probability $\eta$ is 30\% and $k$ is empirically set to 2. For inference, We employ the DDIM\cite{ho2020denoising} scheduler with a sampling step of 35 and set classifier scale $\omega$ to 3.5. 

\begin{figure*}
    \centering
    \includegraphics[width=1\linewidth]{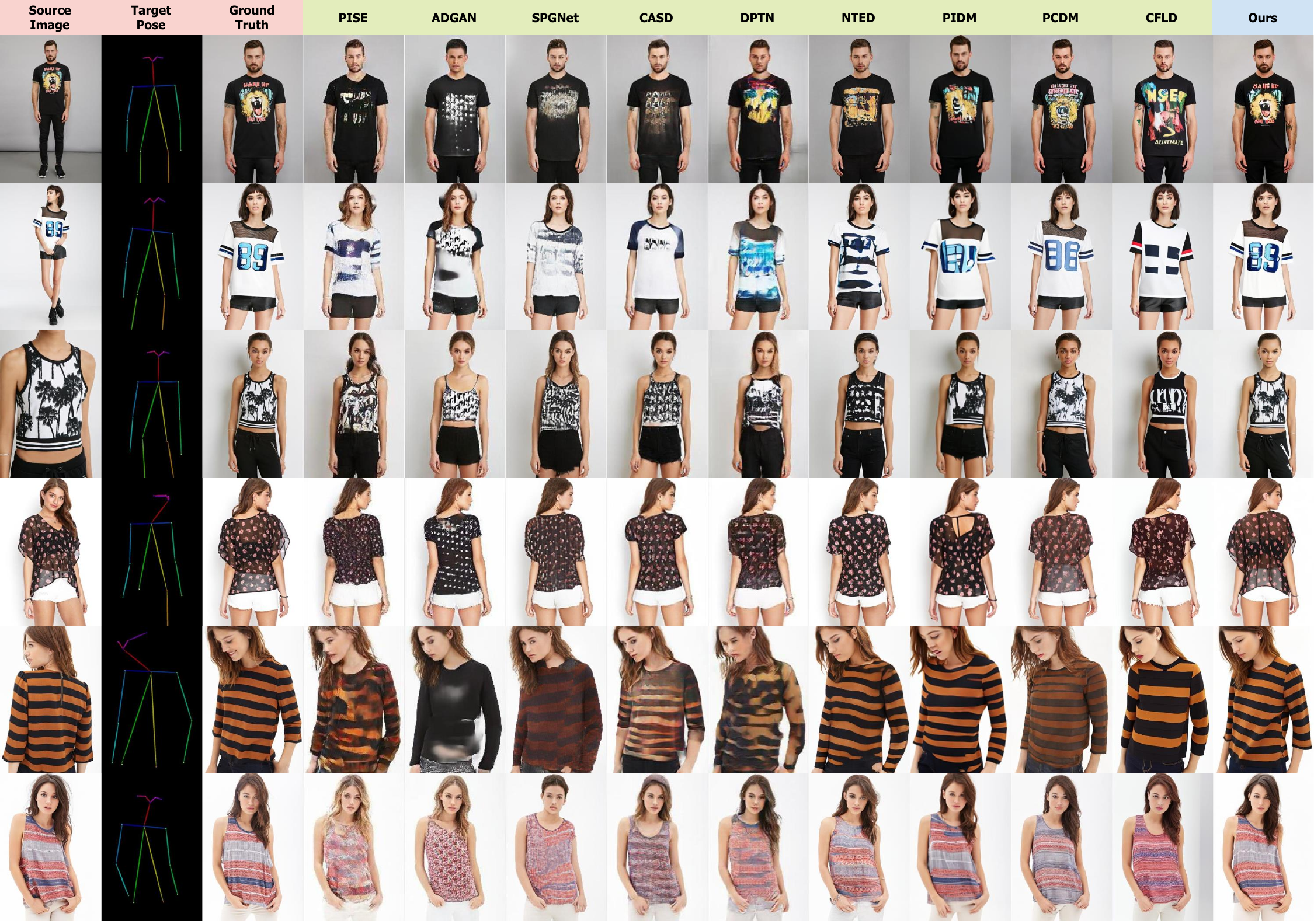}
    \caption{Qualitative comparisons with PISE \cite{zhang2021pise}, ADGAN\cite{men2020controllable}, SPGNet \cite{lv2021learning}, CASD \cite{zhou2022cross}, DPTN \cite{zhang2022exploring}, PIDM \cite{bhunia2023person}, NTED \cite{ren2022neural}, PCDM \cite{shen2024advancing} and CFLD \cite{lu2024coarse}.}
    \label{qualitative_figure}
    \vspace{-10pt}
\end{figure*}

\section{Experiments}
\subsection{Setup}
\noindent \textbf{Dataset.}
We follow \cite{shen2024advancing,lu2024coarse} to conduct extensive experiments on In-Shop Clothes Retrieval benchmark from DeepFashion dataset and evaluated on 512$\times$352 and 256$\times$176 resolutions respectively. 
The dataset is comprised of 52,712 high-resolution images featuring clean backgrounds and diverse fashion models. 
Furthermore, we evaluate the model trained on the DeepFashion dataset using the high-resolution in-the-wild dataset WPose\cite{li2024unihuman}, which consists of 2,304 pairs. This dataset features a more diverse and complex range of backgrounds and human poses.

\noindent \textbf{Objective metrics.}
We use four commonly employed metrics, namely \textit{Structure Similarity Index Measure} (SSIM) \cite{wang2004image}, \textit{Peak Signal to Noise Ratio} (PSNR), \textit{Learned Perceptual Image Patch Similarity} (LPIPS) \cite{zhang2018unreasonable}, and \textit{Fr$\acute{e}$chet Inception Distance} (FID) \cite{heusel2017gans}, to assess quality \& fidelity of generated images. To better assess the authenticity and quality of the generations, we introduce the widely adopted LLM-based metric, Q-Align \cite{wu2023q}, which contains both Image Quality Assessment (IQA) and Image Aesthetic Assessment (IAA).

\noindent \textbf{Subjective metrics.}
We employ three subjective metrics in \cite{bhunia2023person}: R2G \cite{ma2017pose}, G2R \cite{ma2017pose}, and J2b \cite{siarohin2018deformable,bhunia2023person}.
R2G and G2R represent the ratios of real images classified as generated images and generated images classified as real images, respectively. Higher values for R2G and G2R indicate a greater similarity to real images. Meanwhile, J2b reflects user preferences by comparing the proportion of users selecting the best image from a group of images generated by different methods.
\subsection{Quantitative and Qualitative Comparison}
\label{subsec:quantitative and qualitative comparison}
\begin{table}[t!]
    \centering
    \Large
    \caption{Quantitative comparisons with SOTAs on image quality. $^\dagger$We reproduce these results based on the checkpoints instead of generated images provided by the authors. $^\ast$Results are cited from PoCoLD without publicly available checkpoints or generated images. Others are cited from NTED.}
    \label{tab:main_experiment}
    \resizebox{1\linewidth}{!}{
    
\begin{tabular}{l| c| c| c| c| c| c}
    \toprule
         \textbf{Method} & \textbf{Venue} & \textbf{LPIPS$\downarrow$} & \textbf{SSIM$\uparrow$} & \textbf{PSNR$\uparrow$} & \textbf{FID$\downarrow$} & \textbf{IQA $\&$ IAA$\uparrow$}   \\
    \midrule
        \rowcolor{gray!20}
            \multicolumn{7}{l}{\textit{DeepFashion (256 $\times$ 176)}} \\
             PATN\cite{zhu2019progressive} & CVPR 19' & 0.253 & 0.6714 & - & 20.728 & - $\backslash$ - \\
             ADGAN$^\dagger$\cite{men2020controllable} & CVPR 20' & 0.225 & 0.673& 17.010 & 14.540 & 0.250 $\backslash$ 0.582\\
             GFLA\cite{ren2020deep} & CVPR 20' & 0.187 & 0.708& - & 9.827 & - $\backslash$ - \\
             PISE$^\dagger$\cite{zhang2021pise} & CVPR 21' & 0.224 & 0.653& 13.645 & 11.518 & 0.218 $\backslash$ 0.535\\
             SPGNet$^\dagger$\cite{lv2021learning} & CVPR 21' & 0.234 & 0.691& 17.198 & 16.254  & 0.275 $\backslash$ 0.554\\
             CASD$^\dagger$\cite{zhou2022cross} & ECCV 22' & 0.195 & 0.720 & 17.562 & 13.064 & 0.253 $\backslash$ 0.598 \\
             NTED$^\dagger$\cite{ren2022neural} & CVPR 22' & 0.175 & 0.718 & 17.540 & 8.684 & 0.308 $\backslash$ 0.597\\
             DPTN$^\dagger$\cite{zhang2022exploring} & CVPR 22' & 0.200 & 0.710 & 17.032 & 17.526 & 0.158 $\backslash$ 0.379 \\
             PIDM$^\dagger$ \cite{bhunia2023person} & CVPR 23' & 0.186 & 0.676 & 16.185 & \textbf{6.663} & \underline{0.328} $\backslash$ 0.569 \\
             \color{gray}PoCoLD$^\ast$\cite{han2023controllable} & \color{gray}ICCV 23' & \color{gray}0.164 & \color{gray}0.731 & - & \color{gray}8.067 & - $\backslash$ - \\
             PCDM$^\dagger$\cite{shen2024advancing} & ICLR 24' & \underline{0.163} & \underline{0.725} & 17.828 & 7.741 & 0.297 $\backslash$ 0.692\\   
             CFLD$^\dagger$\cite{lu2024coarse} & CVPR 24' & 0.176 & 0.722 & \underline{17.883} & 7.326 & 0.325 $\backslash$ \underline{0.693} \\    
             \textbf{Ours} &  & \textbf{0.151} & \textbf{0.727} & \textbf{18.123} & \underline{7.265} & \textbf{0.344} $\backslash$ \textbf{0.724}\\
    \midrule
        \rowcolor{gray!20}
            \multicolumn{7}{l}{\textit{DeepFashion (512 $\times$ 352)}} \\
             CocosNet2\cite{zhou2021cocosnet} & CVPR 21' & 0.226 & 0.723 & - & 13.325 & - $\backslash$ -\\
             NTED$^\dagger$\cite{ren2022neural} & CVPR 22' & 0.199 & 0.735 & 16.3918 & 7.633 & 0.547 $\backslash$ 0.828 \\
             \color{gray}PoCoLD$^\ast$ \cite{han2023controllable} &  \color{gray}ICCV 23' &  \color{gray}0.192 &  \color{gray}0.743 & - &  \color{gray}8.416 & - $\backslash$ -\\
             PCDM$^\dagger$\cite{shen2024advancing} & ICLR 24' & \underline{0.198} & 0.727 & 17.013 & 7.147 & 0.510 $\backslash$ 0.866 \\ 
             CFLD$^\dagger$\cite{lu2024coarse} & CVPR 24' & 0.201 & \underline{0.734} & \underline{17.101} & \underline{7.102} & \underline{0.554} $\backslash$ \underline{0.884} \\ 
            \textbf{Ours} &  & \textbf{0.180} & \textbf{0.744} & \textbf{17.627} & \textbf{7.05} & \textbf{0.566} $\backslash$ \textbf{0.895} \\
            \color{gray} Ground Truth &  & \color{gray}0.000 & \color{gray} 1.000 & \color{gray} $+\infty$  & \color{gray} 8.028 & \color{gray} 0.571 $\backslash$ \color{gray} 0.947\\
    \midrule
        \rowcolor{gray!20}
            \multicolumn{7}{l}{\textit{WPose (512 $\times$ 352)}} \\
            PCDM$^\dagger$\cite{shen2024advancing} & ICLR 24' & \underline{0.682} & 0.2611 & \underline{10.429} & - & 0.326 $\backslash$ 0.1986  \\
            CFLD$^\dagger$\cite{lu2024coarse} & CVPR 24' & 0.708 & \underline{0.326} & 9.789 & - & \underline{0.470} $\backslash$ \underline{0.271}  \\
            \textbf{Ours} & & \textbf{0.583} & \textbf{0.331} & \textbf{11.457} & - & \textbf{0.565} $\backslash$ \textbf{0.337} \\
            \color{gray} Ground Truth &  & \color{gray}0.000 & \color{gray} 1.000 & \color{gray} $+\infty$ & - & \color{gray} 0.839 $\backslash$ \color{gray} 0.528 \\
    \toprule
\end{tabular}

    }
\end{table}
We conduct a comprehensive comparison with SOTA approaches including 13 methods encompass GAN-based, flow-based, attention-based and diffusion-based,
along with a qualitative evaluation of the latest nine SOTA methods.

\noindent \textbf{Quantitative comparison.}
We conducted a comprehensive quantitative comparison of our method with 13 SOTAs across two datasets, with the results in Table \ref{tab:main_experiment}. Upon analyzing these results, it is clear that our method significantly outperforms existing techniques across nearly all metrics, with the exception of the FID score at the resolution of 256$\times$176. Notably, we achieved substantial improvements in metrics reflecting human preferences, such as LPIPS and LLM-based evaluations, compared to previous GAN-based and diffusion-based methods. This enhancement can be attributed to our proposed model's ability to better capture fine-grained clothing details, resulting in more realistic and higher-quality images.
Furthermore, in the WPose dataset, which features complex backgrounds and poses, our method surpasses the two latest diffusion-based approaches(CFLD,PCDM) across all objective metrics. This finding underscores the superior generalization and robustness of our model in handling diverse and intricate scenarios.
It is noteworthy that although our method has a slightly lower FID than PIDM, this may be attributed to PIDM's overfitting to the dataset distribution, as mentioned in previous studies \cite{lu2024coarse,han2023controllable,shen2024advancing}. 

\noindent \textbf{Qualitative comparison.}
%
In Fig \ref{qualitative_figure}, we compare the visualization our method with nine SOTAs. We observe that:
\ding{182} Previous methods, whether GAN-based or diffusion-based, struggle to retain complex clothing patterns featuring designs or text, even in simple pose transformation scenarios, as demonstrated in \textit{rows 1-2}. In contrast, our method significantly outperforms existing approaches in terms of clothing consistency by incorporating a mask-guided attention fusion mechanism, which effectively captures fine-grained details of garments.
\ding{183} In observing the more extreme pose transformation scenarios in \textit{rows 3}, we attribute our ability to consistently generate images while effectively preserving the clothing details of the source image to the designed local region enhancement and mask-guided overlapping area attention mechanisms. Although the latest diffusion-based methods, such as CFLD and PCDM can generate a rough skeletal representation of the person, they either result in deformations or fail to retain fine-grained patterns.
\ding{184} The last three rows illustrate scenarios where the target pose necessitates the visualization of areas that are not visible in the source image. The results indicate that our method does not overfit and is capable of reasonably generating images of the unseen regions, while also demonstrating superior visual consistency compared to other approaches.

\begin{SCfigure}{}{}
    \includegraphics[width=0.75\linewidth]{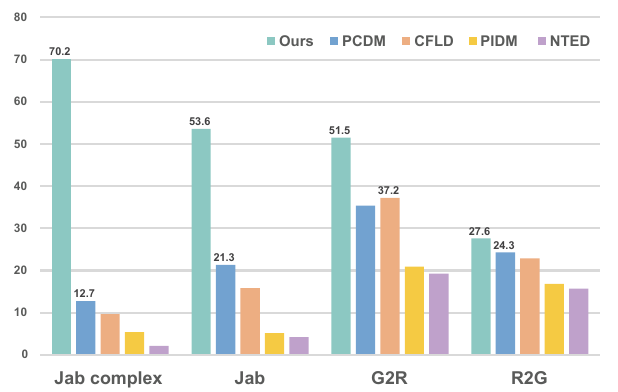}
    \caption{{\color{black}User study results in terms of R2G,G2R and Jab metrics. Higher values in metrics means better quality of generated results.}
    \label{user study}
    }
\end{SCfigure}

\noindent \textbf{User study.}
%
To better illustrate the advantage of our method, we further conduct a user study with 30 computer science volunteers, following the experimental setup of PIDM \cite{bhunia2023person}.
As shown in \figref{user study}, we have the following observations:
\ding{182} For the R2G and G2R metrics, we randomly select 30 images from each generation method and the real dataset to form a test dataset. Volunteers discern whether each image is generated or real. Our method significantly outperforms other methods in G2R metric, with over half of the images perceived as real, while no significant differences are found in R2G metric.
\ding{183}  We also conduct two Jab experiments. In one set, we select 30 images with complex patterns from each method. In the other set, we randomly select 30 images for fair comparison. Our method significantly outperformed others on both sets, achieving scores of 70.2 and 53.6 respectively. These results demonstrate that our method excels in preserving consistency in both complex and normal images while aligning more closely with human aesthetics.
\begin{table}[t]
    \centering
    \caption{Ablation study on different contributions.}
    \label{tab:ablation}
    \resizebox{1\linewidth}{!}{
    \begin{tabular}{l|cccc}
    \toprule
        \makecell[c]{\textbf{Method}} & \textbf{LPIPS$\downarrow$} & \textbf{SSIM$\uparrow$} & \textbf{PSNR$\uparrow$} & \textbf{FID$\downarrow$} \\
    \midrule
             \textbf{Ours} & \textbf{0.1803} & \textbf{0.7444} & \textbf{17.627} & \textbf{7.055}  \\
             B4 (w/o. HPFA) & 0.2077 & 0.7149 & 16.686 & 7.358 \\
             B3 (w/o. CAA) & 0.1874 & 0.7363 & 17.426 & 7.513\\
             B2 (w/. Dual-UNet) & 0.2081 & 0.7059 & 16.267 & 7.647 \\
             B1 (w/. CLIP) & 0.2762 & 0.6895 & 14.613 & 11.982\\
    \bottomrule
\end{tabular}

    }
\end{table}
\begin{SCfigure}
    \centering
    \includegraphics[width=0.85\linewidth]{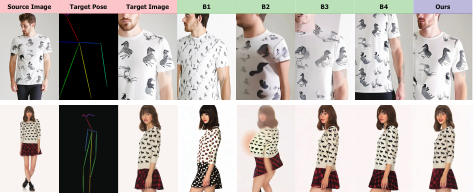}
    \caption{Visualization results of ablation study}
    \label{fig:fig_ablation}
\end{SCfigure}
\subsection{Ablation Study}
\label{subsec: ablation study}

To demonstrate the effectiveness of our contributions, we conduct ablation study on DeepFashion with \tableref{tab:ablation} and \figref{fig:fig_ablation}. 
\textbf{B1} represents our adoption of pre-trained encoders (\ie CLIP) to extract features from the source images like before methods. We design a stacket multi-layer perceptron to map the features to the shape of the $l$-th layer in TargetNet, thereby implementing the fusion mechanism proposed in our method.
\textbf{B2} utilizes the original architecture of \cite{hu2024animate}, and fine-tuning it for 10 epochs on the DeepFashion dataset for a fair comparison.
\textbf{B3} and \textbf{B4} stands for the removal of two proposed attention mechanism.
The results of \textbf{B1} in \secref{sec:limitation} indicate that directly utilizing CLIP fails to preserve clothing details. Furthermore, the naive concatenate fusion like \textbf{B2} results in significant distortion of the characters during large-scale pose transformations. The proposed HPFA module incorporates parsing masks as constraints on attention, enhancing the capability to capture relevant features while minimizing distortions caused by irrelevant regions. Consequently, \textbf{B3} achieves a higher SSIM score compared to \textbf{B2}, indicating improved structural consistency. Due to the presence of certain parsing regions that occupy a small proportion of the overall image or have low attention scores, resulting in poor detail generation, our proposed CAA module adaptively enhances features in local regions to achieve finer-grained generation, as illustrated in \figref{fig:fig_ablation}. In summary, the proposed modules demonstrate complementary effects compared to the baseline, achieving significant improvements in both quantitative and qualitative aspects.

\subsection{Application}
Our method inherits the flexibility of diffusion models, allowing for fashion-related tasks without additional training.

\noindent \textbf{Style Transfer.}
\begin{figure}
    \centering
    \includegraphics[width=1.0\linewidth]{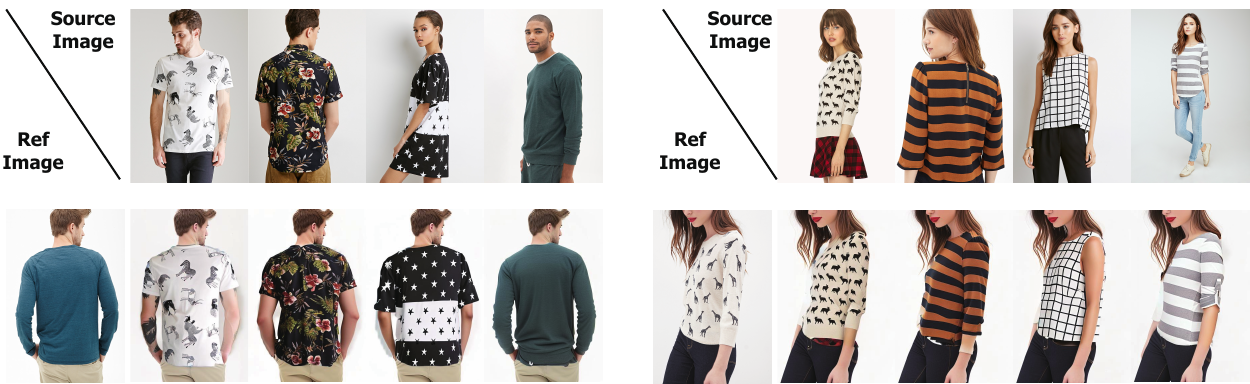}
    \caption{Style transfer results of our method. We can edit the reference image to incorporate pattern features from the source image while ensuring consistency in appearance and pose.}
    \label{fig:fig_styletransfer}
    \vspace{-10pt}
\end{figure}
We select the region of interest from the reference image $\boldsymbol{y}^{r e f}$ to obtain a binary mask $\boldsymbol{m}$. During sampling, the relation $\boldsymbol{y}_t=\boldsymbol{m} \odot \boldsymbol{y}_t+(1-\boldsymbol{m}) \odot \boldsymbol{y}_t^{r e f}$ is applied to obtain the noise at each step $\boldsymbol{t}$, where $\boldsymbol{y}_t$ and $\boldsymbol{y}_t^{r e f}$ represent the predicted noise for the source image and reference image at step $\boldsymbol{t}$ based on the reference image's pose. As shown in Fig \ref{fig:fig_styletransfer}, our method preserves the region of interest in $\boldsymbol{y}^{r e f}$ while generating image details that are visually consistent with the source image.

\section{Conclusion}

In this paper, we introduced a novel human-parsing-aware Siamese network framework for pose-guided person image synthesis (PGPIS) that effectively preserves both facial features and clothing patterns. Our approach comprises three key innovations: the introduction of a dual identical UNet architecture (SourceNet and TargetNet), a human-parsing-guided fusion attention module, and a CLIP-guided attention alignment module. 
Through extensive experiments on the in-shop clothes retrieval benchmark at multiple resolutions, our method significantly outperformed 13 SOTA baselines, successfully generating high-quality images that maintain both facial identity and intricate clothing details - including text patterns, regular textures, and irregular patterns - even under challenging pose transformations. This advancement addresses critical limitations of existing approaches and offers practical value for the retail industry by enabling the efficient generation of realistic product images in various poses while maintaining precise clothing details, thereby reducing the need for additional photo shoots.

{
    \small
    \bibliographystyle{ieeenat_fullname}
    \bibliography{main}
}


\end{document}